\documentclass[a4paper]{article}

\usepackage[english]{babel}
\usepackage[utf8x]{inputenc}
\usepackage[T1]{fontenc}
\usepackage{amssymb}

\usepackage[a4paper,top=3cm,bottom=2cm,left=3cm,right=3cm,marginparwidth=1.75cm]{geometry}

\usepackage{amsmath}
\usepackage{graphicx}
\usepackage[colorinlistoftodos]{todonotes}
\usepackage[colorlinks=true, allcolors=blue]{hyperref}
\usepackage{multirow}

\def\x{{\mathbf x}}
\def\y{{\mathbf y}}
\def\l{{\mathbf l}}
\DeclareMathOperator*{\argmax}{arg\,max}

\title{CTCModel: a Keras Model for Connectionist Temporal Classification}
\author{Yann Soullard, Cyprien Ruffino, Thierry Paquet}

\begin{document}
\maketitle

\begin{abstract}
We report an extension of a Keras Model, called CTCModel, to perform the Connectionist Temporal Classification (CTC) in a transparent way. Combined with Recurrent Neural Networks, the Connectionist Temporal Classification is the reference method for dealing with unsegmented input sequences, i.e. with data that are a couple of observation and label sequences where each label is related to a subset of observation frames. CTCModel makes use of the CTC implementation in the Tensorflow backend for training and predicting sequences of labels using Keras. It consists of three branches made of Keras models: one for training, computing the CTC loss function; one for predicting, providing sequences of labels; and one for evaluating that returns standard metrics for analyzing sequences of predictions. 
\end{abstract}

\section{Introduction}

Recurrent Neural Networks (RNN) are commonly used for dealing with sequential data. The last few years, many developments have been proposed to overcome some limitations which have allowed RNN reaching remarkable performance. For instance, Long Short-Term Memory (LSTM) \cite{hochreiter1997long} and Gated Recurrent Units (GRU) \cite{cho2014properties} mitigate the vanishing (and exploding) gradient problem encountered with Back-Propagation-Through-Time \cite{werbos1990backpropagation}. They also provide a solution for modeling long-range dependencies. Bidirectional systems \cite{graves2005framewise, schuster1997bidirectional} allow to take into account both the left and right context by introducing a forward and backward process. Multi-dimensional Recurrent Neural Networks \cite{graves2009offline} apply recurrent connexions on each dimension which allows accessing a wider contextual information and to deal with input images of variable sizes (both in pixel width and height). In addition, convolutional layers are commonly used to extract features (this relates to an encoder part) that are given in input to a recurrent network \cite{zuo2015convolutional} (a decoder part). More recently, attention models have been proposed with success \cite{bluche2017scan, kim2017joint, song2017end}. 

In many applications on sequential data such as speech and handwritting recognition or activity recognition in videos, a data is an observation sequence $\x=(\x_1,...\x_T)$ of any length T to which one wants to associate a sequence of labels $\y$ of length L, with $L \leqslant T$. During training, a system is trained to model examples in the form of input observation sequence and output labels couples ($\x$, $\y$). The training dataset is incomplete, as the labeling related to each observation $\x_t$ for $1 \leqslant t \leqslant T$ is not known. In others words, the observation sequences are unsegmented as the subset of observation frames related to each label is unknown. For training a recurrent neural network on such data, Alex Graves et al. introduced the Connectionist Temporal Classification (CTC, \cite{graves2006connectionist}). The CTC approach relies on dynamic programming, a Forward-Backward algorithm, to compute a specific loss function based on the probabilities of the possible paths which can produce the output label sequence. 

In the Keras functionnal API, one can define, train and use a neural network using the class Model. The loss functions that can be used in a class Model have only 2 arguments, the ground truth \textit{y\_true} and the prediction \textit{y\_pred} given in output of the neural network. At present, there is no CTC loss proposed in a Keras Model and, to our knowledge, Keras doesn't currently support loss functions with extra parameters, which is the case for a CTC loss that requires sequence lengths for batch training. Indeed, as observation and label sequences are of variable length, one has to provide the length of each sequence (observation and label), which allows to not consider padding in the loss computation. 

In this paper, we present a way for training observation sequences in Keras using the Connectionist Temporal Classification approach. Our proposal extends the Keras Model class to perform a CTC training and decoding. It relies on several Keras Model and on CTC functions in Tensorflow. To the next, we recall the CTC approach (section \ref{sec:introCTC}) and 
describe the model architecture that has been defined (section \ref{sec:CTCModel}). Then, we show how to use CTCModel with Python and present some results on a public dataset in section \ref{sec:expe}. 


\section{Introduction to the CTC approach}
\label{sec:introCTC}

The Connectionist Temporal Classification is a remarkable solution to avoid a pre-segmentation of the training examples and a post-processing to transform the outputs of a Recurrent Neural Network into label sequences. The last layer (softmax layer) of the network contains one unit per label that outputs the probability of having the corresponding label at a particular time step. When one uses the CTC approach, an additional unit is defined to model the probability of having a blank label (i.e. no class label). One denotes $\mathcal{B}$ a function that maps a probability sequence $\pi$ from the network output, i.e. a sequence of probabilities of observing a specific label at a given time frame, into a predicted sequence $\l$, i.e. a sequence of labels of length less than or equal to the one of the input observation sequence. $\mathcal{B}$ consists in removing the repeated labels and then the blank predictions. The conditional probability of having a label sequence $\l$ given an observation sequence $\x$ is the sum of the probabilities of the paths $\pi$ for which $\mathcal{B}(\pi)=\l$:
\begin{equation}
p(\l | \x) = \sum_{\pi \in \mathcal{B}^{-1}(\l)} p(\pi | \x)
\end{equation}

This is computed in an efficient way using a Forward-Backward Algorithm. The decoding task consists in finding the most probable label sequence given an observation sequence:
\begin{equation}
\l^* = \argmax_{\l} p(\l | \x)
\label{eq:decoding}
\end{equation}
In practice, solving equation \ref{eq:decoding} may be time-consuming and may be approximated using a faster method. 
For instance, A. Graves et al. \cite{graves2006connectionist} proposed the \textit{best path decoding} and \textit{prefix search decoding} methods to get an approximate solution. In the best path decoding, finding the most probable path consists in selecting the most probable label for every time frame. On the other hand, the prefix search decoding consists in dividing the output sequence between time steps where the blank label is highly probable and then applying the standard Forward-Backward algorithm on every subset. 

\section{CTCModel: an extension of a Keras Model}
\label{sec:CTCModel}

A Model is the way to manage a neural network in Keras. It has at least two arguments, the network input(s) and the network output(s). The network will include all layers required in the computation of the output given the input. A Keras model contains a number of methods for training the network and testing and evaluating datasets using batches or generators. At compilation time, the model is configured for training: one has to provide at least an optimization method (e.g. 'SGD', 'rmsprop', 'Adam') and a loss function. 

A CTC loss function requires 4 inputs. In addition to the output of the network and the label sequence, the length of each sequence is required in order to limit the loss computation to the sequence lengths. This is crucial for dealing with sequences of variable lengths as sequences are padded to form a tensor of fixed size which is related to a batch. The output of a recurrent network is of equal length than the observation sequence. To our knowledge, Keras doesn't currently support loss functions with extra parameters so it is not possible to use a CTC loss function in a standard way (i.e. by only defining a CTC loss when the model is compiled). A solution to make use of the CTC approach is that the user defines his own loss in a Lambda layer and then instantiates a Model with the custom loss as output. In this way, the loss computation is not performed in a standard manner but directly in the graph computations and a dummy loss has to be given when the model is compiled. A CTC decoding approach can also be defined in a similar way. Based on these solutions, CTCModel has been defined to make use of the Connectionnist Temporal Classification approach in a transparent way. 

\subsection{CTCModel architecture}

CTCModel can be seen as a Keras Model defined to perform the CTC training and decoding steps. At initialization step, it requires the input and output of a recurrent neural network architecture defined in Keras. Let $\x$ be an observation sequence of length $T$ and $\y$ its label sequence of length $L$ where each term $y_l$ is an element $c$ of a set of classes $\mathcal{C}$. Let $\mathcal{\tilde C}$ denotes the set of possible classes (including the blank label), in other words $\mathcal{\tilde C} = \mathcal{C} \bigcup \{b\}$ where $b$ denotes the blank label. The output of the network is a tensor containing the conditional probabilities $p(c|\x_t)$ for each class $c \in \mathcal{C}$ and $1 \leqslant t \leqslant T$. This is commonly the output of a softmax activation layer on a fully connected layer applied at every time frame (e.g. a \textit{TimeDistributed(Dense)} in Keras). This relates to a tensor of dimension \textit{batch\_size $\times$ T $\times$ $|\mathcal{\tilde C}|$}, where $|.|$ denotes the cardinal of a set. If sequences are of variable length, $T$ relates to the length of the longer observation sequence in the batch as it is required to deal with tensors of fixed-sizes. 

\begin{figure}[h]
\centering
\includegraphics[width=.7\textwidth]{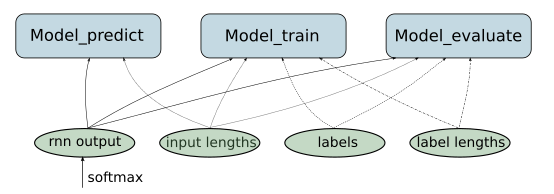}
\caption{Illustration of the CTCModel architecture containing three Model defined in Keras. Model\_train and Model\_evaluate have 4 inputs while Model\_predict has only the 2 inputs related to the observation sequences.}
\label{fig:ctcmodelarchi}
\end{figure}

CTCModel is composed of 3 Keras models: one for training (Model\_train), one for predicting (Model\_predict) and one for evaluating sequences of predictions (Model\_evaluate). It is illustrated in Figure \ref{fig:ctcmodelarchi}. These models are automatically defined when CTCModel is compiled. Each one has a specific loss function (related to the specific task), thus only the optimizer used for training is requested at compile time. 

To the next, we will present more in details the 3 models shown in Figure \ref{fig:ctcmodelarchi}. We will see that CTCModel contains the main functions defined in a Keras model. Some new methods have been defined in the CTCModel implementation while some others have not been implemented again because they are accessible through one of the 3 models as these are Keras Model. 

\subsection{Training, predicting and evaluating}

Inputs and outputs of each Keras model is related to the inputs and outputs of CTC functions defined in Tensorflow. The training model (Model\_train) is based on the \textit{ctc\_batch\_cost} function in Tensorflow. As illustrated in Figure \ref{fig:ctcmodelarchi}, it has 4 inputs: the inputs of the recurrent network, the label sequences, the input lengths and the label lengths. Each one is a Tensor which is initialized at compile time, except the input of the network which is preliminary defined by the user in a standard manner. The standard functions defined in a Keras model for training a neural network are proposed: \textit{train\_on\_batch}, \textit{fit} and \textit{fit\_generator}. 
In case of generators, one returns an input structure x and a label structure y in a standard way but x is a list containing the four inputs highlighted in Figure \ref{fig:ctcmodelarchi} and y is a dummy structure that respects the dimensions of y (y is not used here as the label sequence is given in the input structure x). 

The predicting model (Model\_predict) relies on the Tensorflow function \textit{ctc\_decode}. It requires only two inputs: the inputs of the network (i.e. the Input layer in Keras) and the input lengths. Decoding parameters (greedy, beam\_width, top\_paths) proposed in the Tensorflow function \textit{ctc\_decode} can be defined when CTCModel is instantiated (by default, a fast best-path search is applied). Similarly to the training model, standard functions to predict sequences of labels are defined for various applications: \textit{predict\_on\_batch}, \textit{predict} and \textit{predict\_generator}. In the case of a generator, the outputs of the generator are the same as for the training model (i.e. there are 4 elements in x) and only the input observations and input lenghts are considered. 

Finally, the evaluating model (Model\_evaluate) is based on the \textit{edit\_distance} function defined in Tensorflow. Three metrics are proposed and can be computed: the loss ('loss'), the label error rate ('ler') and the sequence error rate ('ser'). They are defined in the \textit{metrics} argument. Whatever the metric, the 4 inputs defined above are required. The loss and sequence error rate are computed on the entire input dataset while the label error rate is returned for every sequence (this is a list of label error rate values). 

\subsection{Additional functionalities}

In addition to the evaluation functions, CTCModel contains functions that return only the loss (\textit{get\_loss\_on\_batch}, \textit{get\_loss} and \textit{get\_loss\_generator}). It returns the likelihood on the entire dataset and for each observation sequence. Similarly, the probabilities used in input of the CTC approach can be obtained using \textit{get\_probas\_on\_batch}, \textit{get\_probas} or \textit{get\_probas\_generator}. It returns a list of probability matrix of size T $\times$ number of labels (+1 with the blank which is the last probability), where T is the length of the sequence (one recalls that data can be of different lengths). 

Besides, CTCModel proposes two methods for saving and loading models. In those methods, each Keras model (i.e. Model\_train, Model\_predict, Model\_evaluate) is loaded or saved in a standard way with json. Note that the function \textit{save\_model} allows to save the network architecture, not the weights of the model. The weights can be saved using the callback \textit{ModelCheckpoint} in Keras. The \textit{load\_model} function allows to load the weights that have been saved if a file name containing the weights is given. Besides, a pickle file is also created to save and load the model hyper-parameters. 

\section{Experimentations}
\label{sec:expe}

We now present how to use CTCModel in Python and then show some results we obtained using the public French dataset RIMES. 

\subsection{The use of CTCModel}

\begin{figure}[h]
\centering
\includegraphics[width=.7\textwidth]{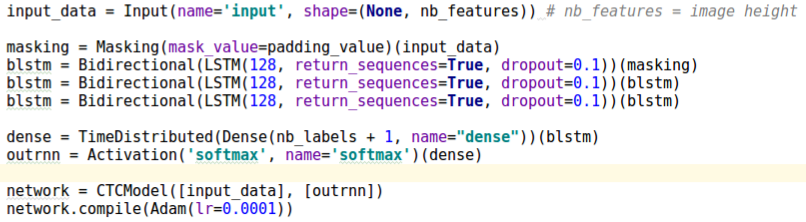}
\caption{Example of a code for using the CTCModel model in Keras.}
\label{fig:define_ctcmodel}
\end{figure}

CTCModel is defined in a standard way as specified with Keras Model. First, one has to define a recurrent network. As illustrated in Figure \ref{fig:define_ctcmodel}, we first define an Input layer where both batch size and observation length are unknown, while the number of features, e.g. height of text line images or Histogram of Oriented Gradients features for example, has to be specified. Then, one defines the recurrent layers and the output layer. In Figure \ref{fig:define_ctcmodel} there are three bidirectional LSTM layers with 128 outputs and then a Dense layer with a softmax activation. The number of outputs is the number of labels + 1 (one output per label + the blank label). 

Then we instantiate a CTCModel where inputs and outputs are respectively the Tensors given by the Input layer and the one given by the softmax activation layer. Then, the network is compiled: notice that the only argument to provide is the optimization method (here Adam with a learning rate of 1e-4), as the only possible loss is the CTC loss function. In contrast to a standard Keras Model, evaluation metrics are specified as argument of the evaluate method, not at compilation time. 

\begin{figure}[h]
\centering
\includegraphics[width=.7\textwidth]{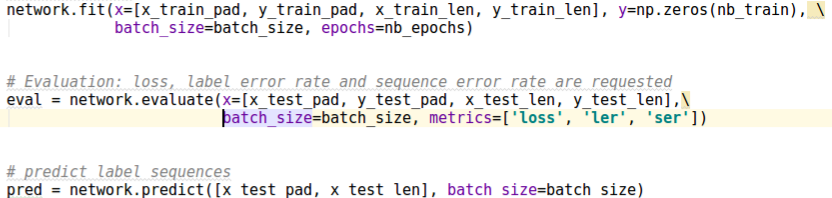}
\caption{Example of the use of methods fit, evaluate and predict, proposed in CTCModel.}
\label{fig:use_ctcmodel}
\end{figure}

For the use of CTCModel methods, one recalls that inputs x and y are defined in a particular way as x contains the input observations, the labels, the input lengths and the label lengths while y is a dummy structure. Thus, the \textit{fit} and \textit{evaluate} methods require the specific inputs x, while the \textit{predict} function only requires the observation sequences and observation lengths as input, as illustrated in Figures \ref{fig:ctcmodelarchi} and \ref{fig:use_ctcmodel}. Note that input observations and label sequences have to be padded to get input tensors of fixed size in input of the Keras and Tensorflow functions. 

\subsection{Results on the RIMES dataset}

We conducted some experiments on the RIMES dataset \cite{grosicki2008rimes} composed of French handwritten documents. We applied a standard recurrent neural network composed of 4 bi-directional layers with 100 and 200 units per direction on line images rescaled to 100 pixels height. Line images are standardized (i.e. 0 mean and unit variance) and given in input of the model. This setting leads to train less than one million parameters and 3.4 million for the two respective models. Then, we experimented a convolutional recurrent neural network with a model architecture containing 8 convolutional layers with max-pooling and dropout after two successive convolutional layers. Here, there are 32 (x2), 64 (x2), 128 (x2) and 256 (x2) filters respectively of size 3x3 and then a fully-connected layer with 256 units. To perform convolution layers on input images of fixed-size, one applies a sliding window procedure with a stride of 4 on the line images (one recalls that they are of variable lengths). In this way, one built a sequence of sub-images that are of dimensions 32 $\times$ 64, which relates to 32 time frames and 64 features per time-frame (i.e. normalized pixels). Then, there are 2 bi-directional layers with 256 units per direction. This setting leads to train less than 6 million of parameters. Table \ref{tab:perfRIMES} shows the Character Error Rate (CER) we obtained at convergence. Note that these results are obtained only using the Optical Character Recognition system (without any language model). 

\begin{table}[h]
\centering
\resizebox{0.7\textwidth}{!}{%
\begin{tabular}{|c|c|c|c|}
\hline
Model & Train set & Validation set & Test set \\ \hline
RNN 100 + CTCModel & $6.3\%$ & $11.2\%$ & $12.2\%$   \\ \hline
RNN 200 + CTCModel & $2.6\%$ & $9.3\%$ & $10.6\%$   \\ \hline
conv8 RNN + CTCModel & $0.1\%$ & $6.8\%$ & $7.6\%$ \\ \hline
\end{tabular}%
}
\caption{Character Error Rate on the RIMES dataset.}
\label{tab:perfRIMES}
\end{table}

\section{Conclusion}

We present CTCModel, an extension of a Keras Model to perform the Connectionist Temporal Classification approach. It relies on efficient methods defined in Tensorflow for training, by computing the CTC loss, and predicting, by performing a CTC decoding. The main Keras Model methods have been proposed in CTCModel and can be used in a standard way. The main difference with a standard Keras Model is the specific input structure containing both the observation sequences, the input observation lengths, the label sequences and the label lengths. Two evaluation metrics, the label error rate and the sequence error rate can also be computed in a transparent way from the CTC decoding. 

\section*{Acknowledgment}

This work has been supported by the French National grant ANR 16-LCV2-0004-01 “Labcom INKS”. This work is founded by the French region Normandy and the European Union. Europe acts in Normandy with the European Regional Development Fund (ERDF).

\bibliographystyle{plain}
\bibliography{my_bib}

\end{document}